\documentclass{isprs} 
\usepackage{subfigure}
\usepackage{setspace}
\usepackage{geometry} 
\usepackage{epstopdf}
\usepackage[labelsep=period]{caption}  
\usepackage[british]{babel} 
\usepackage[hang]{footmisc}

\begin{document}

\title{A Novel Solution for Drone Photogrammetry with Low-overlap Aerial Images using Monocular Depth Estimation}
\date{}


\author{Jiageng Zhong\textsuperscript{1,\dag}, Qi Zhou\textsuperscript{2,\dag}, Ming Li\textsuperscript{1,3,*}, Armin Gruen\textsuperscript{3},  Xuan Liao\textsuperscript{4}}

\address{
		\textsuperscript{1 }State Key Laboratory of Information Engineering in Surveying Mapping and Remote Sensing, Wuhan University,\\ Wuhan 430079, China - lisouming@whu.edu.cn\\
		\textsuperscript{2 }School of Remote Sensing Information Engineering, Wuhan University, Wuhan 430079, China\\
		\textsuperscript{3 }Institute of Geodesy and Photogrammetry, ETH Zurich, Zurich, 8039, Switzerland\\
		\textsuperscript{4 }Department of Land Surveying and Geo-Informatics, The Hong Kong Polytechnic University, Hong Kong, China\\
}



\abstract{
Low-overlap aerial imagery poses significant challenges to traditional photogrammetric methods, which rely heavily on high image overlap to produce accurate and complete mapping products. In this study, we propose a novel workflow based on monocular depth estimation to address the limitations of conventional techniques. Our method leverages tie points obtained from aerial triangulation to establish a relationship between monocular depth and metric depth, thus transforming the original depth map into a metric depth map, enabling the generation of dense depth information and the comprehensive reconstruction of the scene. For the experiments, a high-overlap drone dataset containing 296 images is processed using Metashape to generate depth maps and DSMs as ground truth. Subsequently, we create a low-overlap dataset by selecting 20 images for experimental evaluation. Results demonstrate that while the recovered depth maps and resulting DSMs achieve meter-level accuracy, they provide significantly better completeness compared to traditional methods, particularly in regions covered by single images. This study showcases the potential of monocular depth estimation in low-overlap aerial photogrammetry.
}


\maketitle


\renewcommand{\thefootnote}{} 
\footnotetext[1]{* Corresponding author}
\footnotetext[2]{\dag \hspace{.05em} Equal technical contribution}

\section{Introduction}\label{Introduction}
 
\begin{figure}
 	\centering
 	\includegraphics[width=0.8\columnwidth]{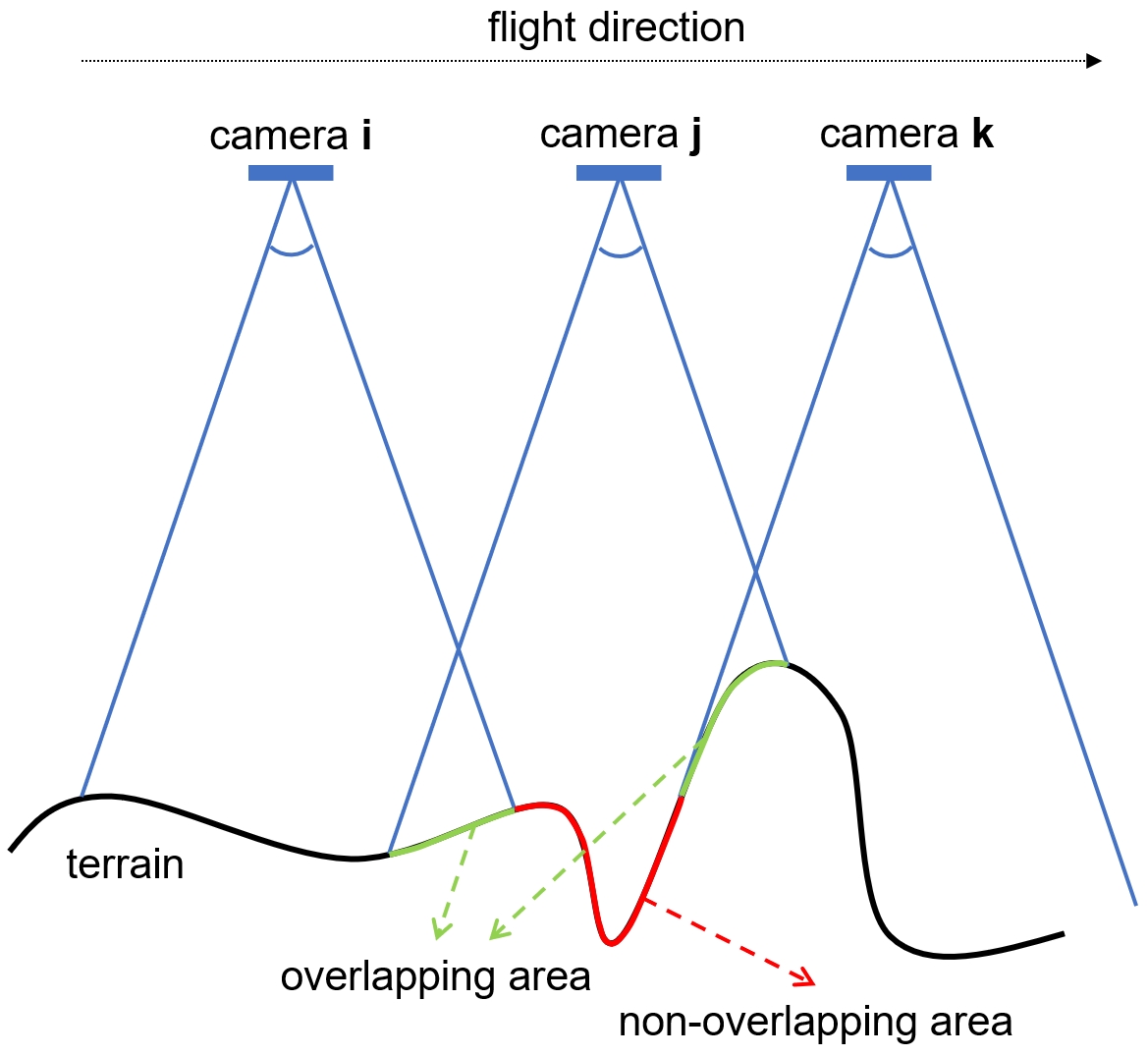}
 	\caption{\raggedright Low altitude, high speed, and significant ground variation can all lead to low aerial image overlap.}
 	\label{fig1}
 \end{figure}

 Drone photogrammetry has become widely adopted in urban planning and environmental monitoring due to its efficiency and versatility. Classical photogrammetry derives 3D information by calculating ray intersections within overlapping image regions to create geospatial products, such as 3D terrain maps, Digital Surface Models (DSMs), and true orthophotos. This methodology typically requires a high degree of image overlap, usually ranging from 60\% and 80\%, to ensure reliable 3D reconstruction through binocular or multi-view stereo matching methods \cite{schonberger2016pixelwise,zhou2022digital} . However, certain critical applications, such as disaster assessment, emergency response, and aerial reconnaissance, impose strict temporal constraints while simultaneously requiring high-resolution imagery. These conditions often necessitate drone operations at low altitudes and high velocities \cite{liu2017novel}. As illustrated in Figure 1, these operational parameters frequently compromise image overlap, sometimes reducing it to merely 10–30\%. The challenge is further exacerbated in regions with significant terrain variation, where valid image overlap may fall well below the standard requirements for aerial surveys. This type of imagery is referred to as low-overlap aerial imagery. The technical challenges presented by low-overlap scenarios are particularly demanding. In contrast to conventional photogrammetry, where overlapping images facilitate 3D coordinate determination through ray intersection, low-overlap conditions result in areas being captured only once. These non-overlapping regions, illustrated by the red curve segment in Figure \ref{fig1}, present a fundamental limitation: without sufficient overlap, ray intersections cannot be established, preventing stereo measurements. This not only limits the generation of digital surface models (DSMs) but also prevents the creation of complete orthomosaic maps, which are essential for many geospatial applications.
 
 \begin{figure*}
 	\centering
 	\includegraphics[width=1.98\columnwidth]{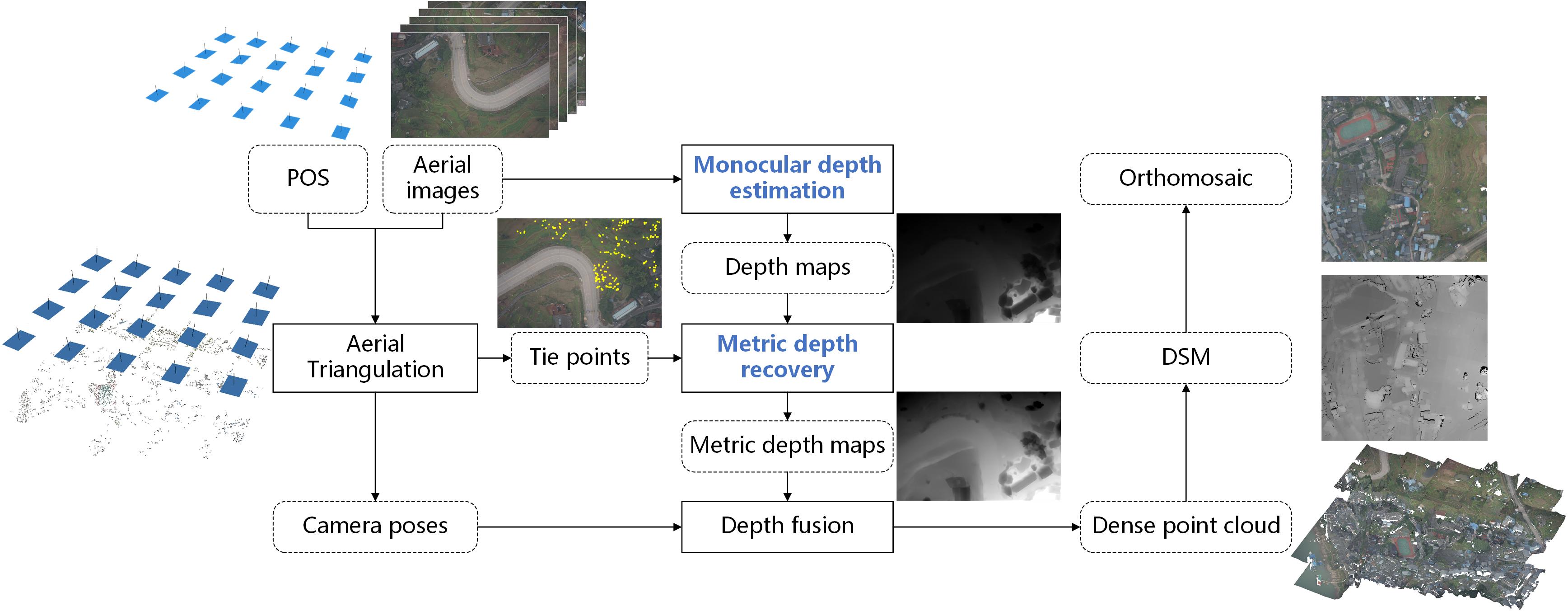}
 	\caption{The aerial photogrammetry solution suitable for low-overlap images.}
 	\label{fig2}
 \end{figure*}
 
 The fundamental challenges posed by low-overlap imagery necessitate an innovative approach to 3D reconstruction, particularly in non-overlapping areas where traditional ray intersection methods fail. A promising solution emerges from monocular depth estimation techniques, which can predict the depth of each pixel directly from a single image, eliminating the reliance on overlapping images for 3D reconstruction. In recent years, rapid advancements in computer vision and deep learning have driven significant progress in monocular depth estimation, resulting in the development of robust and versatile approaches. For instance, MiDaS \cite{ranftl2020towards,ranftl2021vision,birkl2023midas} leverage large-scale datasets with 3D supervision, addressing scale differences across datasets to enable zero-shot testing. Depth Anything \cite{yang2024depth,yang2024depth2} demonstrates outstanding performance across diverse scenarios through an advanced unsupervised learning paradigm. Metric3D \cite{yin2023metric3d,hu2024metric3d} incorporates intrinsic parameters to transform images into a canonical camera space, effectively addressing scale inconsistencies. These advanced methods leverage deep neural networks trained on vast datasets to understand complex spatial relationships and scene geometry, enabling them to infer depth values with impressive accuracy even in challenging conditions. By leveraging such state-of-the-art techniques, it becomes feasible to estimate depth in non-overlapping regions, paving the way for complete and dense 3D terrain reconstruction even under conditions of sparse or minimal image overlap. However, while monocular depth estimation holds great promise, there remain critical challenges to its direct application in geospatial analysis. A key limitation is the scale ambiguity in most monocular depth estimation methods. Even models specifically designed for metric depth estimation, such as Metric3D, may struggle to accommodate diverse camera models and application contexts. This issue is critical, as accurate metric-scale depth is essential for producing geospatial products, such as DSMs and orthomosaic maps.
  
 In this study, we propose a novel framework to address incomplete reconstruction in aerial surveys with low-overlap imagery, building upon aerial triangulation and image orientation. Our approach leverages state-of-the-art monocular depth estimation to generate dense depth maps for each aerial image, effectively addressing coverage gaps in non-overlapping regions. To overcome the scale ambiguity inherent in monocular predictions, we utilize sparse but highly accurate depth values derived from tie points generated during aerial triangulation for metric depth recovery. Through fusion of these recovered depth maps, our approach produces comprehensive geospatial products, including high-resolution dense point clouds, DSMs, and orthomosaic maps that ensure complete coverage of the surveyed area.

 \section{Methodology}
 
 This study takes drone imagery with POS (Position and Orientation System) information as an example, which is a common type of drone data, and proposes an aerial photogrammetry method suitable for low-overlap aerial images as shown in Figure \ref{fig2}. The method consists of the following steps. Initially, aerial triangulation is performed on the aerial images with POS information to determine the accurate poses of the images. At the same time, tie points for each image are obtained. Tie points refer to keypoints identified to establish geometric relationships between the images. Subsequently, a monocular depth estimation model is employed to generate depth maps for each aerial image, though these initial depth values exhibit significant deviations from ground truth. These depth values are then recovered to metric depth with tie points. Using the poses obtained from triangulation, the recovered depth maps are converted into point clouds and fused to produce a dense point cloud. Finally, the dense point cloud is rasterized into a Digital Surface Model (DSM), which is used for orthorectification to generate the true orthomosaic map.

 Among these steps, the most critical factors affecting the quality of the final product are monocular depth estimation and depth recovery. In the monocular depth estimation process, deep learning models are applied using color aerial images as inputs. In some cases, camera parameters are used as auxiliary data. The direct output is typically a disparity map where disparity values are inversely proportional to depth. A high-performance model is expected to restore the relative depth relationships across different areas of the captured scene. Depth recovery uses tie points as references to convert depth values with metric ambiguity into accurate metric depth values. After aerial triangulation, a set of tie points is obtained for each image along with their image plane coordinates and their coordinates in the photogrammetric coordinate system. Combined with camera parameters, the monocular depth estimation value $z_{mono}$  for each tie point and its true depth value $z_{gt}$  can be determined. A function $f$ is fitted to establish the relationship between $z_{mono}$ and $z_{gt}$, expressed as $z_{gt}=f(z_{mono})$. Preliminary experiments indicate that a first-order rational function model achieves acceptable results with low computational complexity. The depth recovery model can thus be expressed as:
 \begin{equation}\label{eq1}
 	 z_{gt}=f(z_{mono})=\frac{a \cdot z_{mono} + b}{c \cdot z_{mono} + d} ,  
 \end{equation}
 where $a$, $b$, $c$, and $d$ represent the rational function coefficients, which can be estimated through nonlinear least squares curve fitting.
 
  \section{Experimental Results and Analysis}
 
 \subsection{Survey Area and Data}
 
\begin{figure}
 	\centering
 	\includegraphics[width=0.98\columnwidth]{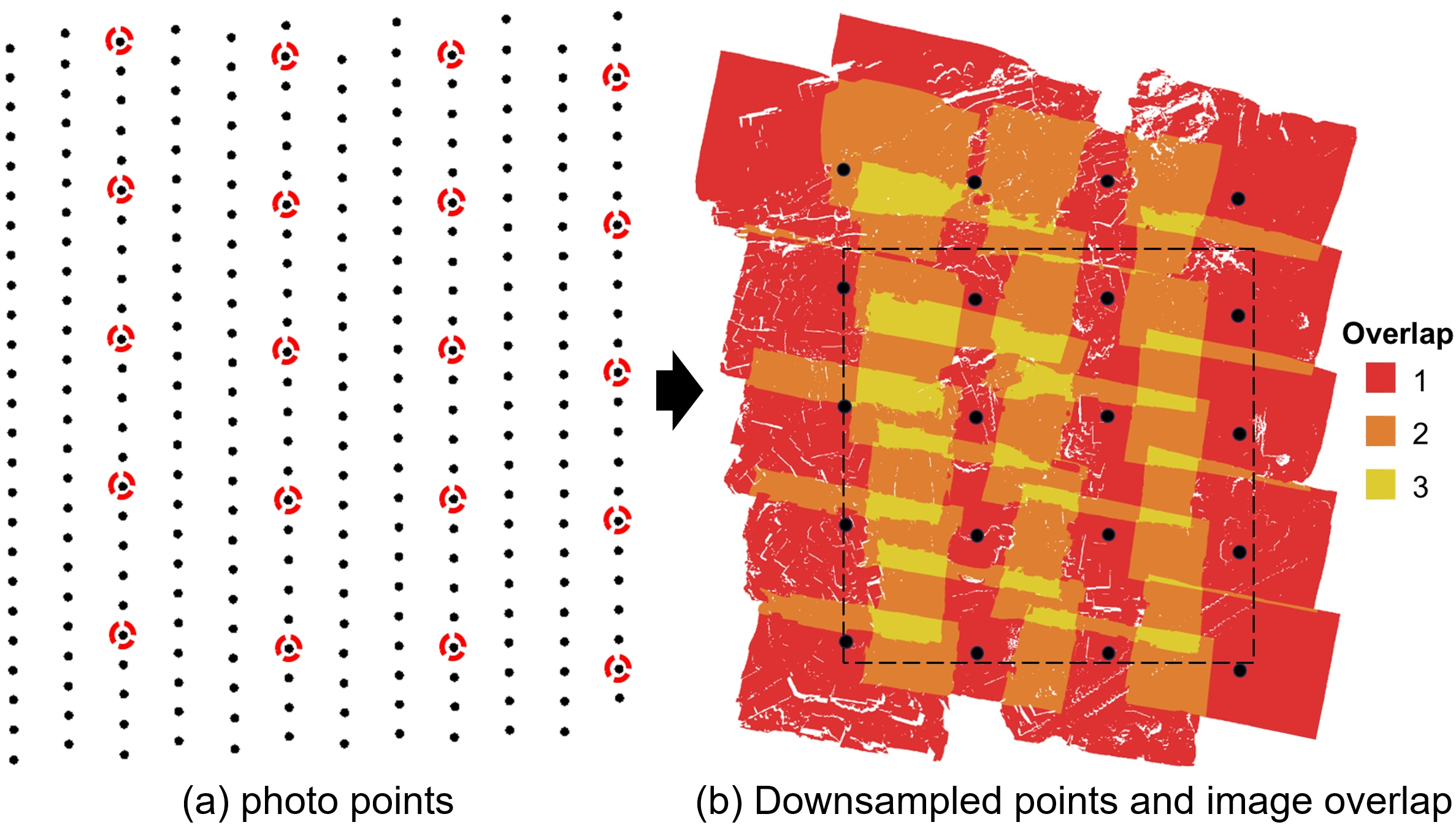}
 	\caption{\raggedright Data overview. (a) Black dots represent photo points from the original dataset, with red circles marking the selected low-overlap images after downsampling. (b) Distribution and overlap pattern of the downsampled images; the black rectangle indicates the validation area, measuring 400×400 meters.}
 	\label{fig3}
 \end{figure}

 The data for this study was collected in Chongqing, China, a mountainous urban area with elevations ranging from 200 to 300 meters and a terrain characterized by moderate undulations. The images were captured using an unmanned aerial vehicle flying at an altitude of approximately 200 meters, equipped with a camera capable of capturing images at a resolution of 6000 × 4000 pixels. The original dataset was acquired using strip-wise flight patterns with high overlap ratios to generate high-precision, high-resolution products. It includes 296 images, as shown in Figure \ref{fig3}(a), accompanied by POS data. The forward overlap is approximately 80\%, while the side overlap is around 70\%. Ground control points were established within the survey area to validate aerial triangulation accuracy, which achieves centimeter-level precision. This level of precision guarantees that the resulting depth maps, digital surface models and orthomosaic maps are of high quality and suitable for use as ground truth in subsequent experiments. The dataset is then systematically downsampled to create a low-overlap image set, as depicted in Figure \ref{fig3}(b). This subset consists of 20 images with forward and side overlaps reduced to approximately 20\% and 40\%, respectively. In this configuration, large portions of the survey region are covered by only a single image, highlighted in red.
 
 \subsection{Monocular Depth Estimation and Recovery}
 
 This section evaluates the performance of monocular depth estimation models and their effectiveness in depth recovery. The monocular depth estimation models used in this study include MiDaS \cite{ranftl2020towards,ranftl2021vision,birkl2023midas} and Depth Anything \cite{yang2024depth,yang2024depth2}. These methods have released multiple versions over the past few years, each offering various model weights. For this study, the latest versions of these models are selected, along with the weights demonstrating optimal overall performance. Specifically, MiDaS v3.1 \cite{birkl2023midas} is utilized with the BEiT transformer backbone \cite{bao2021beit} trained at high resolution, while Depth Anything v2 \cite{yang2024depth2} is employed using weights based on the DINOv2 large model \cite{oquab2023dinov2}. 
 
\begin{figure*}
 	\centering
 	\includegraphics[width=1.98\columnwidth]{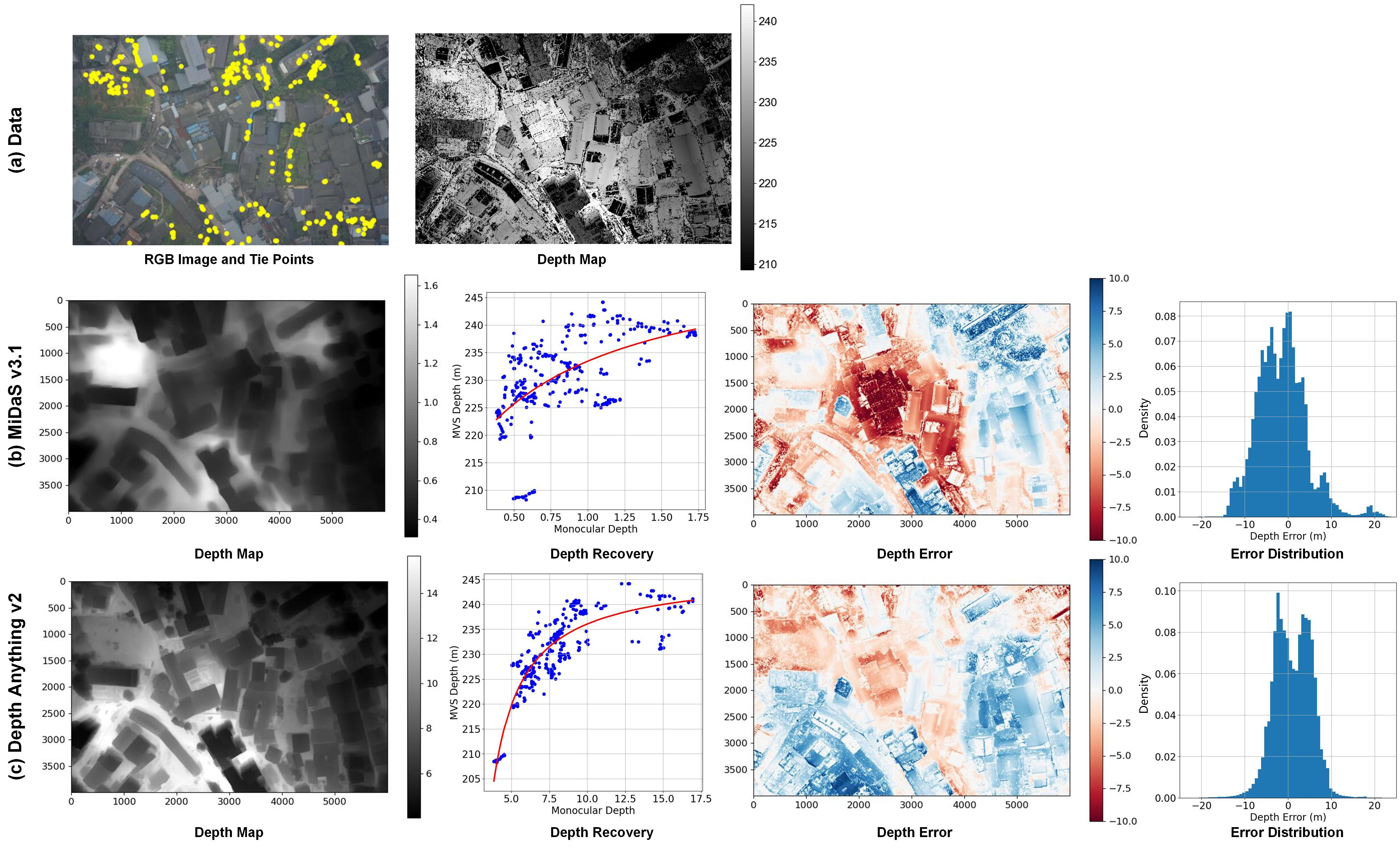}
 	\caption{\raggedright Depth recovery results for depth maps generated by different monocular depth estimation models. (a) The input aerial RGB image and the corresponding depth map generated by Metashape. Yellow points on the RGB image indicate tie points derived from aerial triangulation. (b) Results from MiDaS v3.1. (c) Results from Depth Anything v2.}
 	\label{fig4}
 \end{figure*}

\begin{figure*}
 	\centering
 	\includegraphics[width=1.98\columnwidth]{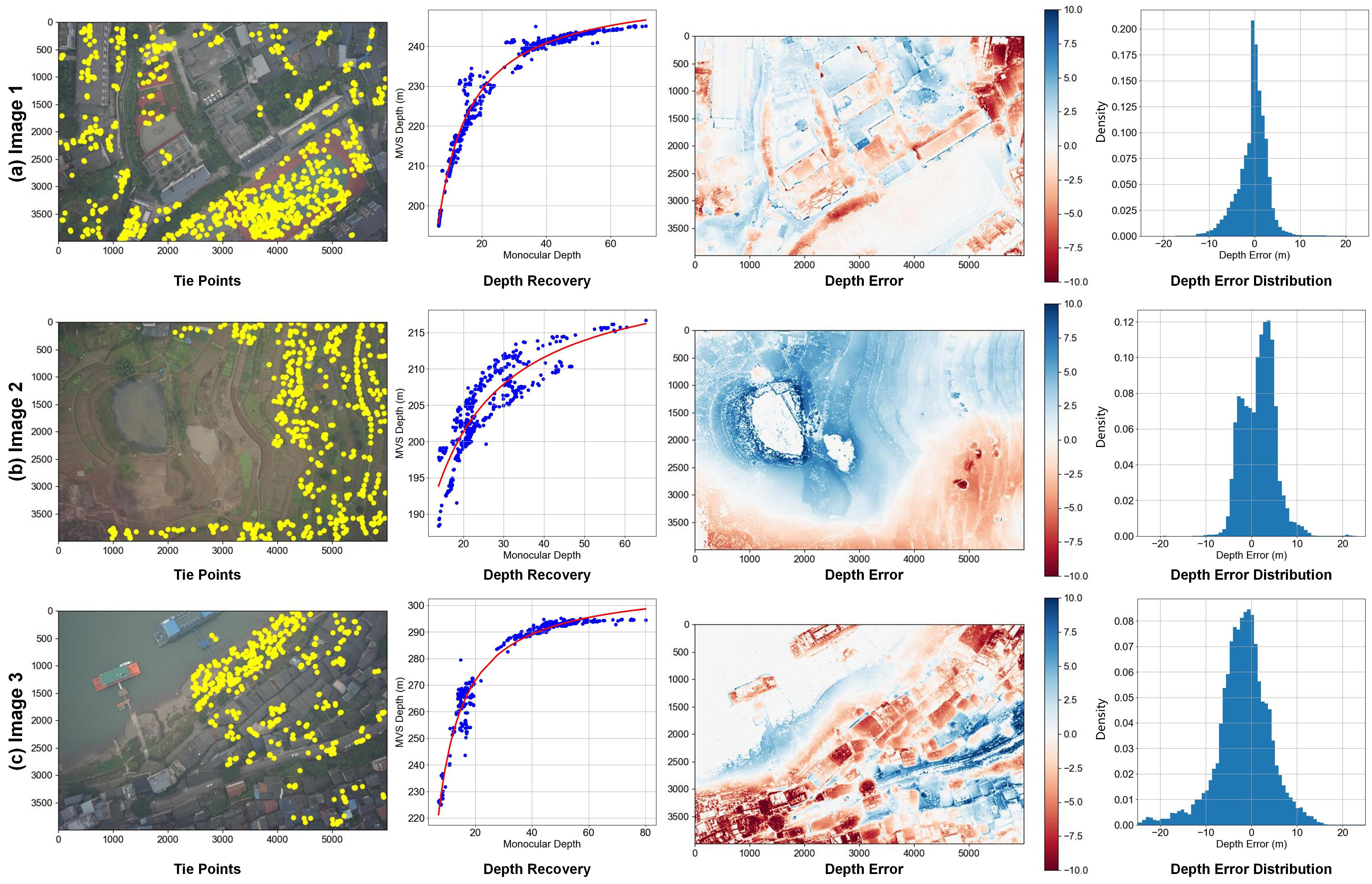}
 	\caption{Additional Results of Depth Anything v2.}
 	\label{fig5}
 \end{figure*}

 To evaluate accuracy, we process the original dataset of 296 images using Metashape, employing Multi-View Stereo (MVS) techniques to generate depth maps for each image as ground truth. For each image, the monocular depth values corresponding to its tie points are fitted to the ground truth depth values using Formula \ref{eq1}. The fitted parameters are then applied to transform the original monocular depth map into metric depth, yielding recovered depth maps. The Root Mean Square Error (RMSE) between the recovered depth map and the ground truth depth map is computed. For the low-overlap dataset, the mean RMSE across all processed images is calculated and referred to as mRMSE. The results for different monocular depth estimation models are presented in Table \ref{tab1}. Both monocular depth estimation networks exhibited meter-level errors, with MiDaS v3.1 demonstrating notably higher errors compared to Depth Anything v2. Figure \ref{fig4} illustrates an example of these results. Visual inspection reveals that MiDaS v3.1 generates depth maps with less detail compared to Depth Anything v2. The latter produces sharper results, especially around building edges. Regarding depth estimation, the depth recovery results clearly demonstrate that Depth Anything v2 achieves better fitting with ground truth values, indicating more consistent estimation across varying depth regions. The depth error visualization further shows that MiDaS v3.1's errors are predominantly concentrated in the central image region, where it struggles to estimate the depth of buildings accurately. Based on these findings, subsequent product creation is conducted using Depth Anything v2. Figure \ref{fig5} presents additional results of Depth Anything v2. In flat areas, such as the playground shown in Figure \ref{fig5}(a), the errors are minimal. However, in regions with significant elevation variation, the errors become more pronounced. This is especially evident in areas with abrupt height changes, such as the terrain in Figure \ref{fig5}(c), where the elevation difference exceeds 50 meters. Similarly, structures like buildings and trees, which lack distinct height features, pose significant challenges for monocular depth estimation from nadir-view imagery.
 
 \begin{table}[ht]
 	\caption{\raggedright The mRMSE results of different monocular depth estimation networks after depth recovery.}
 	\centering
 	\label{tab1}
	\renewcommand\arraystretch{1.1}
 	\begin{tabular}{ll}
 		\hline
 		Model & mRMSE (m)\\
 		\hline
 		MiDaS v3.1 & 8.92 \\
 		Depth Anything v2 & 5.77 \\
 		\hline
 	\end{tabular}
 \end{table} 
 
 \subsection{Photogrammetric Products}
 
 \begin{figure*}
 	\centering
 	\includegraphics[width=1.98\columnwidth]{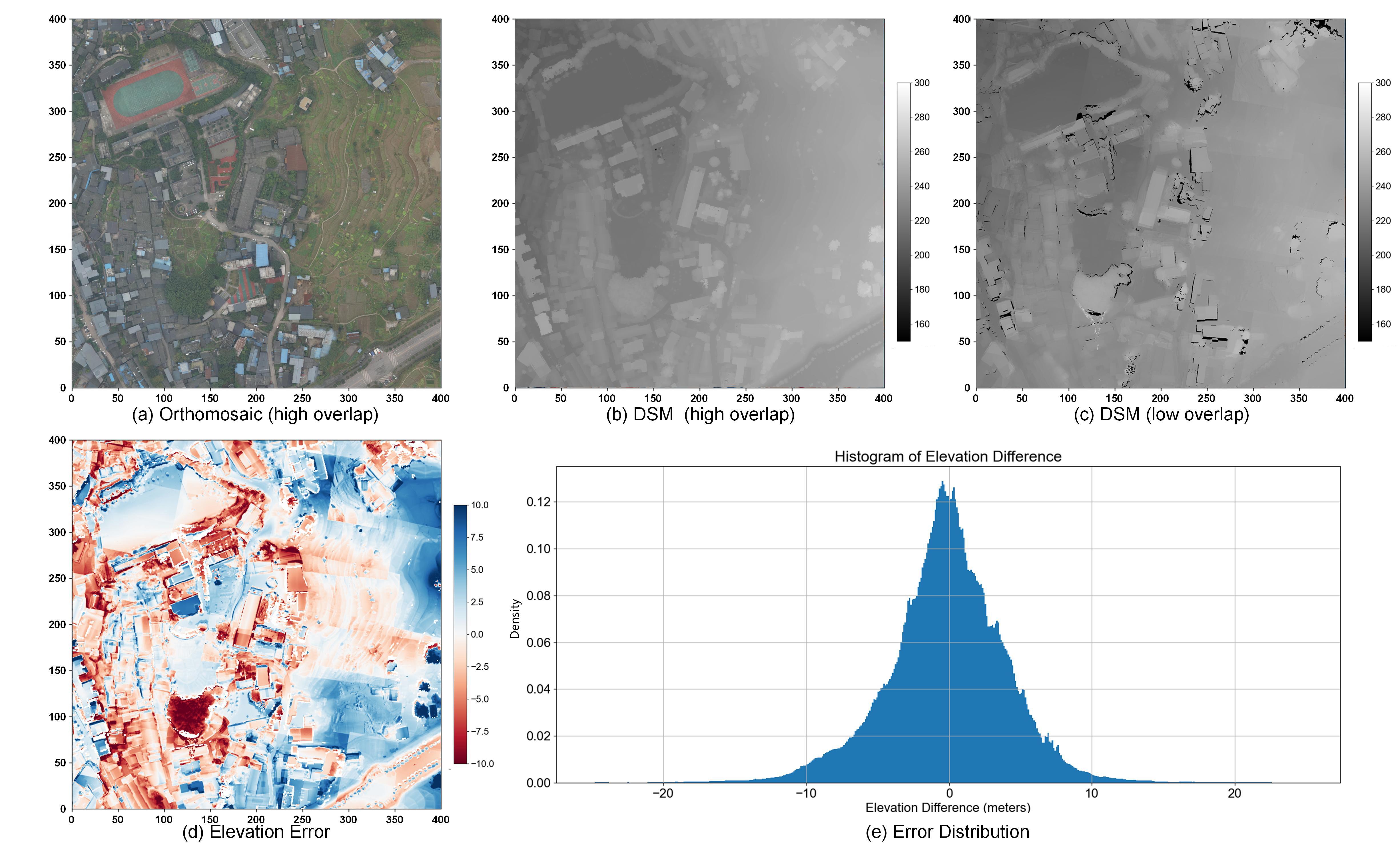}
 	\caption{\raggedright DSM Error Analysis. (a) and (b) represent the orthomosaic and DSM generated from the high-overlap original dataset processed with Metashape, respectively. (c) shows the DSM produced from the low-overlap images using our proposed method. (d) illustrates the error distribution of the DSM in (c), and (e) presents the frequency histogram of the errors.}
 	\label{fig6}
 \end{figure*}
 
 \begin{figure}
 	\centering
 	\includegraphics[width=0.64\columnwidth]{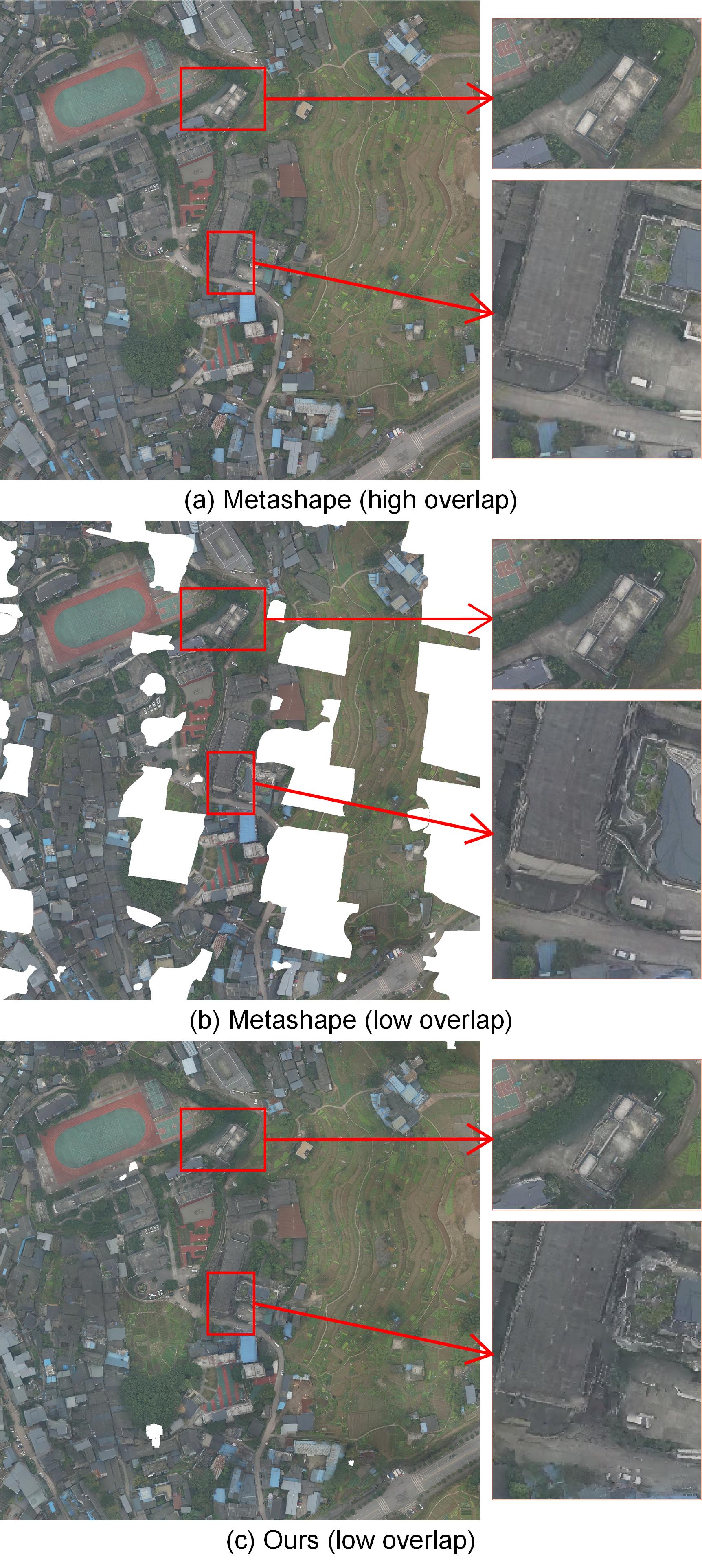}
 	\caption{\raggedright Orthomosaics generated by different settings and approaches.}
 	\label{fig7}
 \end{figure}
 
 Using the recovered depth maps from Depth Anything v2, the DSM and orthomosaic can be generated, as illustrated in Figure \ref{fig6}. Statistical results reveals that the mean absolute error of the elevation data is 4.29 meters, with a median absolute error of 2.40 meters. The histogram in Figure \ref{fig6}(e) indicates that the majority of errors are within 10 meters. Given a flight height of 200 meters, the relative error in most areas is within 5\%. The error visualization in Figure \ref{fig6}(d) shows that a forested area in the lower-left central region exhibits significant errors, exceeding 10 meters. According to the DSM in Figure 6(b), the trees in this area are approximately 15–20 meters tall, indicating that the monocular depth estimation significantly overestimated the height (or underestimated the depth) Similarly, the trees near the edge of the playground in the upper-central area also show considerable errors, suggesting that Depth Anything v2 struggles with depth estimation of trees from a nadir perspective. For buildings, errors are predominantly concentrated along their edges, while roof surfaces show lower errors due to their relative flatness. The terrain in the right section, despite its undulations, exhibits relatively small errors due to its gradual variations. Notable discontinuities in the visualization result from inconsistencies in the overlapping areas of merged dense point clouds. This artifact occurs because, in this experiment, the depth maps are individually converted into point clouds and then fused directly.
  
 After generating DSMs, orthomosaics can be produced, as illustrated in Figure \ref{fig7}. For comparison, we process both the high-overlap original dataset and the downsampled low-overlap dataset using Metashape. It is evident that conventional methods fail to create orthophotos in areas covered by only a single image. Moreover, the neighboring regions exhibit significant errors, resulting in noticeable texture distortions in the final orthomosaic. In contrast, our method successfully creates a nearly complete orthomosaic, demonstrating greater completeness and accuracy compared to conventional methods, although some deficiencies remain at the edges of buildings.

 \section{Conclusion}
 
 In this study, we address the challenge of generating complete photogrammetric products from low-overlap aerial images, where traditional photogrammetric methods often fail. To overcome this limitation, we propose an approach based on monocular depth estimation, leveraging tie points obtained from aerial triangulation to recover metric depth from scale-ambiguous monocular depth maps. This enables comprehensive scene reconstruction and the creation of DSMs and orthomosaics. Experimental results demonstrate that the proposed method effectively generates dense depth information in regions without overlapping images. Under flight conditions at a height of 200 meters, our proposed approach achieves DSM accuracy at the meter level. These findings highlight the promising potential of the method, and we believe this study can serve as a starting point to inspire further research and innovation in this field.

{
	\begin{spacing}{1.17}
		\normalsize
		\bibliography{ISPRSguidelines_authors} 

\begin{thebibliography}{xx}

\bibitem[Bao et al., 2021]{bao2021beit}
Bao, H., Dong, L., Piao, S., Wei, F., 2021.
 Beit: Bert pre-training of image transformers.
 {\em arXiv preprint arXiv:2106.08254}.

\bibitem[Birkl et al., 2023]{birkl2023midas}
Birkl, R., Wofk, D., M{\"u}ller, M., 2023.
 Midas v3. 1--a model zoo for robust monocular relative depth estimation.
 {\em arXiv preprint arXiv:2307.14460}.

\bibitem[Hu et al., 2024]{hu2024metric3d}
Hu, M., Yin, W., Zhang, C., Cai, Z., Long, X., Chen, H., Wang, K., Yu, G., Shen, C., Shen, S., 2024.
 Metric3D v2: A Versatile Monocular Geometric Foundation Model for Zero-shot Metric Depth and Surface Normal Estimation.
 {\em arXiv preprint arXiv:2404.15506}.

\bibitem[Liu et al., 2017]{liu2017novel}
Liu, J., Gong, J., Guo, B., Zhang, W., 2017.
 A novel adjustment model for mosaicking low-overlap sweeping images.
 {\em IEEE Transactions on Geoscience and Remote Sensing}, 55(7), 4089--4097.

\bibitem[Oquab et al., 2023]{oquab2023dinov2}
Oquab, M., Darcet, T., Moutakanni, T., Vo, H., Szafraniec, M., Khalidov, V., Fernandez, P., Haziza, D., Massa, F., El-Nouby, A. et~al., 2023.
 Dinov2: Learning robust visual features without supervision.
 {\em arXiv preprint arXiv:2304.07193}.

\bibitem[Ranftl et al., 2021]{ranftl2021vision}
Ranftl, R., Bochkovskiy, A., Koltun, V., 2021.
 Vision transformers for dense prediction.
 \emph{Proceedings of the IEEE/CVF international conference on computer vision}, 12179--12188.

\bibitem[Ranftl et al., 2020]{ranftl2020towards}
Ranftl, R., Lasinger, K., Hafner, D., Schindler, K., Koltun, V., 2020.
 Towards robust monocular depth estimation: Mixing datasets for zero-shot cross-dataset transfer.
 {\em IEEE transactions on pattern analysis and machine intelligence}, 44(3), 1623--1637.

\bibitem[Sch{\"o}nberger et al., 2016]{schonberger2016pixelwise}
Sch{\"o}nberger, J.~L., Zheng, E., Frahm, J.-M., Pollefeys, M., 2016.
 Pixelwise view selection for unstructured multi-view stereo.
 \emph{Computer Vision--ECCV 2016: 14th European Conference, Amsterdam, The Netherlands, October 11-14, 2016, Proceedings, Part III 14}, Springer, 501--518.

\bibitem[Yang et al., 2024a]{yang2024depth}
Yang, L., Kang, B., Huang, Z., Xu, X., Feng, J., Zhao, H., 2024a.
 Depth anything: Unleashing the power of large-scale unlabeled data.
 \emph{Proceedings of the IEEE/CVF Conference on Computer Vision and Pattern Recognition}, 10371--10381.

\bibitem[Yang et al., 2024b]{yang2024depth2}
Yang, L., Kang, B., Huang, Z., Zhao, Z., Xu, X., Feng, J., Zhao, H., 2024b.
 Depth Anything V2.
 {\em arXiv preprint arXiv:2406.09414}.

\bibitem[Yin et al., 2023]{yin2023metric3d}
Yin, W., Zhang, C., Chen, H., Cai, Z., Yu, G., Wang, K., Chen, X., Shen, C., 2023.
 Metric3d: Towards zero-shot metric 3d prediction from a single image.
 \emph{Proceedings of the IEEE/CVF International Conference on Computer Vision}, 9043--9053.

\bibitem[Zhou et al., 2022]{zhou2022digital}
Zhou, Q., Duan, Y., Zhao, X., Dong, J., Zhang, H., Cao, H., 2022.
 Digital surface model generation from aerial imagery using bridge probability relaxation matching.
 {\em Journal of Applied Remote Sensing}, 16(4), 046511--046511.

\end{thebibliography}
	\end{spacing}
}

\end{document}